# Dealing with uncertainty on the initial state of a Petri net

Iman JARKASS*, Michèle ROMBAUT**
*HEUDIASYC-UTC, Université de Technologie de Compiègne, B.P. 20529, 60205 Compiègne cedex, France
Tel: +33 (0)3.44.23.44.23, E-mail: Iman.Jarkass@utc.fr
**LM2S-UTT, IUT de Troyes, Université de Technologie de Troyes, B.P. 2060, 10010 Troyes cedex, France
Tel: +33 (0)3.25.71.56.88, E-mail: rombaut@univ-troyes.fr

## Abstract

This paper proposes a method to find the actual state of a complex dynamic system from information coming from the sensors on the system himself, or on its environment.
The nominal evolution of the system is a priori known and can be modeled (by an expert, for example), by different methods. In this paper, the *Petri nets* have been chosen. Contrary to the usual use of the Petri nets, the initial state of the system is unknown. So a degree of *belief* is bound to each places, or set of places. The theory used to model this uncertainty is the *Dempster-Shafer's* one which is well adapted to this type of problems.
From the given Petri net characterizing the nominal evolution of the dynamic system, and from the observation inputs, the proposed method allows to determine according to the reliability of the model and the inputs, the state of the system at any time.

## 1 Introduction

A method is proposed to find the state of a complex dynamic system, using information coming from the sensors on the system itself, or on its environment. Such a system is, for instance the driving situations of a vehicle. The aim is to recognize the phase of a particular manoeuvre at any time by using information coming from the proprioceptive sensors of the vehicle such as accelerometers, steering sensor, and so on. The phase is characterized by a temporal sequence of inputs.

The nominal evolution of the system is at priori known and can be modeled (by an expert, for example), by different methods such as Markof's model or Petri nets. In the first part of this paper, it is proposed to give a reminder on the Petri net theory.

Contrary to the usual use of the Petri nets, the initial state of the system is unknown. But the sequence of the events on the receptivities gives more and more information, in order to determine the actual state of the system. So a degree of *belief* is bound to each place, or set of places. The theory used to model this uncertainty is the *Dempster-Shafer's*, the one which is well adapted to this type of problems. In the second part, this theory applied to the Petri nets is detailed.

From the given Petri net characterizing the nominal evolution of the dynamic system, and from the observation inputs, the proposed method allows to determine with the great **reliability** according to the reliability of the model and the inputs, the state of the system at any time. The third part of this paper proposes the global resolution for a sequential Petri net, and for a net with a conflict. Some examples are also given.

## 2 The theory of the Petri nets

The Petri net is a tool for the representation and modeling of dynamic systems [1], [2]. For example, in the intelligent vehicle area [3], the observed vehicle is making a particular manoeuvre. The goal is to determine the phase of the manoeuvre at any time, with just the information of the proprioceptive sensors. In this case, each place of the Petri net can represent a phase that corresponds to a part of the circuit, and each transition represents the threshold on the sensor that model a phase change. A token in a particular place means the vehicle is in the corresponding phase.

## 2.1 The Petri net: definition and graphic presentation

A Petri net (**RdP**) is composed of two knot types: the *places* and the *transitions* shown in figure (1). A **place** is represented by a circle and a **transition** is represented by a line between two places. The number of places and transitions is finished and not null.

A place represents the state of the system, (for instance, the phase of the manoeuvre) and a transition is a relationship connecting two places. It is associated to an event whose occurrence leads to the evolution of the system (for instance, a speed change).

A Petri net is defined by :
$$RdP = <P, T, Pre, Post>$$
where:
$P$: the set of places $P_i$
$T$: the set of transitions $t_j$
$Pre$: $P \times T \longrightarrow \mathbb{N}$. It is the *input* incidence function.
$Post$: $T \times P \longrightarrow \mathbb{N}$. It is the *output* incidence function.

The *Pre* incidence function describes the oriented arcs connecting the places to the transitions. It represents, for each transition $t_j$, the part of the state in which the system is before the state change when $t_j$ occurs. $Pre(P_i, t_j)$ is the weight of the arc $(P_i, t_j)$. The absence of one arc between a place $P_i$ and a transition $t_j$ is denoted by $Pre(P_i, t_j) = 0$.

The *Post* incidence function describes the oriented arcs connecting transitions to the places. It represents, for each transition $t_j$, the part of the state in which the system will be after the occurrence of the state change corresponding to $t_j$. $Post(t_j, P_i)$ is the weight of the arc $(t_j, P_i)$. The absence of an arc between a transition $t_j$ and a place $P_i$ is denoted by $Post(t_j, P_i) = 0$.

In this paper, it is assumed that *Pre* and *Post* take value in $\{0, 1\}$.

The figure (1) represents a Petri net (RdP) with 3 places and 3 transitions. The sets of places and of transitions are :
$P = \{P_1, P_2, P_3\}$
$T = \{t_1, t_2, t_3\}$

The two matrix, *Pre* and *Post*, associated to the *Pre* and *Post* incidence functions are:

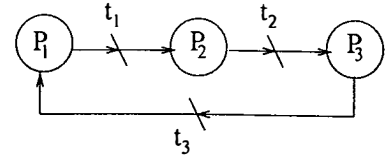

Figure 1: Petri net with 3 places

$$Pre = \begin{matrix} t_1 & t_2 & t_3 \\ \begin{bmatrix} 1 & 0 & 0 \\ 0 & 1 & 0 \\ 0 & 0 & 1 \end{bmatrix} & & \begin{matrix} P_1 \\ P_2 \\ P_3 \end{matrix} \end{matrix}$$

$$Post = \begin{matrix} t_1 & t_2 & t_3 \\ \begin{bmatrix} 0 & 0 & 1 \\ 1 & 0 & 0 \\ 0 & 1 & 0 \end{bmatrix} & & \begin{matrix} P_1 \\ P_2 \\ P_3 \end{matrix} \end{matrix}$$

To each transition is associated a **receptivity** [1] corresponding to an event on the input. Its value is one when the associated event is true.

The vector $R_{(k)}$ is the vector of receptivity $r_j$ at the time $(k)$. ($r_j$: the receptivity of the transition $t_j$).

$$R_{(k)} = [r_{j(k)}]^T, j = 1 \ldots N$$

## 2.2 The dynamic aspect of the Petri net

The dynamic of the system is represented by the motion of a token in the net. To each place of the net a mark is associated. This mark is one when the token is in the place, and zero otherwise. The mark one in a place means the system is in the phase associated to this place. The token moves when the receptivities of the transitions are changed.

The mark of a place $P_i$ at the time $(k)$ is given by $M(P_i)_{(k)}$, $M_{(k)}$ is the associated mark vector.

$$M_{(k)} = [M(P_i)_{(k)}]^T, i = 1 \ldots N$$

From an initial marking $M_{(k)}$, the **firing** of the transitions gives a new mark vector $M_{(k+1)}$ such as :

$$M_{(k+1)} = M_{(k)} - Pre.R_{(k+1)} + Post.R_{(k+1)} \quad (1)$$

The difference of the two matrix $Post - Pre$ has the following property: the sum of the elements on each column is equal to zero which is a fundamental property of the Petri nets. The Petri nets are covered by positive p-invariant, which is the number of tokens in the set. From an initial marking, the next marking of the net can evolve by a **firing** of transitions. Whatever is the evolution, the number of tokens in the net stays constant.

---

[1] A receptivity is a logical proposal, for example, "the speed of the vehicle is hight".

In the particular cases studied here, the value of this number is one, so, at any time:

$$M(P_1) + M(P_2) + \ldots + M(P_n) = 1 \quad (2)$$

In fact, the mark can represent the belief of the fact that the token is in a particular place. If the mark is one, it is sure to be in the corresponding place, and if it is zero, it is sure that not. So, the mark can be seen such as the probability $\mathcal{P}(P_i)$, to be in the corresponding place $P_i$.
So,

$$M(P_i) \equiv \mathcal{P}(P_i)$$
$$\sum_{i=1\ldots N} M(P_i) = \sum_{i=1\ldots N} \mathcal{P}(P_i) = 1 \quad (3)$$

This interpretation of the mark is compatible to the probability theory.

The first studied case concerns a **simple** Petri net, that means a sequential evolution of the net. The second study examines the case of a conflict [2].

The classic Petri net describes perfectly the real evolution of the system, the initial state is well-known, and the events associated to the receptivity of the transitions are certain. In fact, for the complex systems, the net describes approximatively the system (macro states, fuzzy limits between the states), the inputs are not always sure, and the initial state is sometimes unknown [4].

This paper deals with the uncertainty of the initial state parameters. It is assumed that the net is well known, and the inputs are certain. In this case, it is impossible to directly used the Petri net as presented before, because the vector $M$ at time $k = 0$ cannot be initialized. With the classic Petri net, it is impossible to model the uncertainty. In the next section, the uncertainty is modeled by using the theory of evidence (Dempster-Shafer's theory).

## 3 Fundamental notions on the theory of evidence

The theory of evidence, introduced by Dempster in 1960, was rewrited in a more mathematical form by Shafer in 1976 [5]. This theory is known under several names; the theory of beliefs, the theory of plausibilities or again the Dempster-Shafer's theory. It allows to represent the uncertainty on the different hypotheses of a frame of discernment $\Omega$.

This theory comes from the well-known theory of the probability. In this last theory, the uncertainty about several hypotheses of $\Omega$ is modeled by an equal repartition of the probability on them. In the theory of evidence, the probability is given to the union set of them. In fact, this theory allows to define a distribution of probability, named function of **mass** on each element of $2^\Omega$, the set of all parts of $\Omega$. The mass represents the probability to be in a place or a set of places.

In the case of the Petri net, the definition of the frame of discernment $\Omega$ corresponds to the set of hypotheses about the state of the system:

$$\Omega = \{P_1, P_2, P_3, \ldots, P_n\}$$

The set $2^\Omega$ is deduced and has $(2^n - 1)^3$ elements. For the Petri net of n places, $2^\Omega$ is:

$$2^\Omega = \{\{P_1\}, \{P_2\}, \ldots, \{P_1, P_2\}, \ldots, \{P_1, P_2, \ldots, P_n\}\}$$

The **mass m** allocated to each element is therefore defined by:
$$\mathbf{m} : 2^\Omega \longrightarrow [0, 1]$$

such as:

- $\mathrm{m}(\emptyset) = 0$,
- $\sum_{X \subseteq 2^\Omega} \mathrm{m}(X) = 1$

This last property is similar to the conservation of the sum of the mark (equation 2) in classic Petri net.
The mark of the set $\{P_1, P_2, \ldots, P_i\}$ means the mass allocated to this set and it's noted by $M_{\{1,2,\ldots,i\}}$.

The uncertainty about the initial state is modeled by giving all the probability on the set $\{P_1, P_2, \ldots, P_n\}$, and all the other masses equal to zero; this proposition means that it is sure to be in $\Omega$:

$$m(\{P_1, P_2, \ldots, P_n\}) = M_{\{1,2,\ldots,n\}} = 1$$
$$m(X) = 0, \forall X \neq \{P_1, P_2, \ldots, P_n\}$$

In the first paragraph, the vector of the marking is defined on places of the net, it corresponds to the belief to be in a place. In the same manner, a new marking vector is defined with the distribution of masses on $2^\Omega$. Its elements correspond to the belief to be in a set of place. This vector, noted $\mathcal{M}$, is such as,

$$\mathcal{M} = [M_{\{1\}}, M_{\{2\}}, \ldots, M_{\{1,2\}}, \ldots, M_{\{1,2,3,\ldots,n\}}]$$

where $M_{\{i,\ldots,j\}}$ is the mass of belief of the element $X = \{P_i, \ldots, P_j\}$. It is now necessary to determine the evolution of the vector $\mathcal{M}$, when the events come on the transitions.

---

[2] The conflict is defined when at least 2 transitions come from one place.

[3] The st $\emptyset$ is not included in $2^\Omega$

## 4  Global resolution of the system

This section aims to describe the global evolution of the marking vector $\mathcal{M}_{(k+1)}$ at time $(k+1)$, taking into account the vector $\mathcal{M}_{(k)}$ and the inputs $R_{(k+1)}$ at time $(k+1)$, and based on the logical equations of the Petri net. The receptivities $r_j$ take the value in $\{0,1\}$. That leads that the mark $M_X$ of the subset of places $X$ of $2^\Omega$ takes the logical values in $\{0,1\}$.

*In the first step,*
for each element $X$ of $2^\Omega$ and for each possible combination of the receptivities of $R$, noted by $comb$, the element $Y$ of $2^\Omega$, named *transformation* of $X$ by the combination $comb$ of $R$, is determined. $Y$ verifies the following implication:

$$(M_X = 1 \text{ and } comb) \Longrightarrow (M_Y = 1)$$

If $M_{X(k)} = 1$ at time $(k)$, all the places $P_i \in X$ are possible. At the time $(k+1)$ and with a combination of receptivities of $R_{(k+1)}$, all the places $P_j \in Y$ that verify the equation (1) are possible. The places $P_j$ are the transformation of the places $P_i$ by $R_{(k+1)}$. So, $Y$, the *transformation* of $X$ by the combination $comb$ of $R_{(k+1)}$, is the set of the $P_j$.

*In the second step,*
for each $Y$, all the couple of $(X, comb)$ that lead to $Y$ are determined. The new state of $Y$ is a combination of all possible $(X, comb)$.

### 4.1  Example: a simple Petri net with 3 places

For the example of the Petri net presented in figure (1), the frame of discernment is $\Omega = \{P_1, P_2, P_3\}$, and the set $2^\Omega$ is:

$$2^\Omega = \{\{P_1\}, \{P_2\}, \{P_3\}, \{P_1, P_2\}, \{P_1, P_3\},$$
$$\{P_2, P_3\}, \{P_1, P_2, P_3\}\}$$

The vector $\mathcal{M}$ is:

$$\mathcal{M} = [M_{\{1\}}, M_{\{2\}}, M_{\{3\}},$$
$$M_{\{1,2\}}, M_{\{1,3\}}, M_{\{2,3\}}, M_{\{1,2,3\}}]$$

The transformation of each element of $2^\Omega$ is computed for each possible value $comb$ of the receptivity vector of the transitions. A transition $t_i$ has the receptivity $r_i$, which can be 1 or 0, if 1, $r_i$ is true, and if 0, $\bar{r}_j$ is true. $r_i$ means that the transition $t_i$ can be fired, and $\bar{r}_j$ means the transition cannot be fired. For example, the sequence $\bar{r}_1 r_2 \bar{r}_3$, that corresponds to $R = [0,1,0]$, means that only the transition $t_2$ can be fired, if $P_1$ is true.

*First step*
Because there are three transitions, $2^3 = 8$ combinations of the receptivities $r_i$ are possible. The transformation computation of two elements $X$ of $2^\Omega$ is presented as follows:

$$\{P_1\} \begin{cases} \xrightarrow{r_1 \bar{r}_2 \bar{r}_3} \{P_2\} \\ \xrightarrow{\bar{r}_1 r_2 \bar{r}_3} \{P_1\} \\ \xrightarrow{\bar{r}_1 \bar{r}_2 r_3} \{P_1\} \\ \xrightarrow{r_1 r_2 \bar{r}_3} \{P_2\} \\ \xrightarrow{r_1 \bar{r}_2 r_3} \{P_2\} \\ \xrightarrow{\bar{r}_1 r_2 r_3} \{P_1\} \\ \xrightarrow{r_1 r_2 r_3} \{P_2\} \\ \xrightarrow{\bar{r}_1 \bar{r}_2 \bar{r}_3} \{P_1\} \end{cases} \quad \{P_1, P_3\} \begin{cases} \xrightarrow{r_1 \bar{r}_2 \bar{r}_3} \{P_2, P_3\} \\ \xrightarrow{\bar{r}_1 r_2 \bar{r}_3} \{P_1, P_3\} \\ \xrightarrow{\bar{r}_1 \bar{r}_2 r_3} \{P_1\} \\ \xrightarrow{r_1 r_2 \bar{r}_3} \{P_2, P_3\} \\ \xrightarrow{r_1 \bar{r}_2 r_3} \{P_1, P_2\} \\ \xrightarrow{\bar{r}_1 r_2 r_3} \{P_1\} \\ \xrightarrow{r_1 r_2 r_3} \{P_1, P_2\} \\ \xrightarrow{\bar{r}_1 \bar{r}_2 \bar{r}_3} \{P_1, P_3\} \end{cases}$$

This computation is made for all the $X$ elements of $2^\Omega$.

*Second step*
From these maps, inverted transformation is made: for all $Y$ of $2^\Omega$, the corresponding $X$ are determined.

For example, all couples with transformation as $\{P_1\}$ are:

| | | | | |
|---|---|---|---|---|
| $\{P_1\}$ | by | $[\bar{r}_1, \bar{r}_2, \bar{r}_3]$ | $\rightarrow$ | $\{P_1\}$ |
| $\{P_1\}$ | by | $[\bar{r}_1, r_2, \bar{r}_3]^T$ | $\rightarrow$ | $\{P_1\}$ |
| $\{P_1\}$ | by | $[\bar{r}_1, \bar{r}_2, r_3]^T$ | $\rightarrow$ | $\{P_1\}$ |
| $\{P_1\}$ | by | $[\bar{r}_1, r_2, r_3]^T$ | $\rightarrow$ | $\{P_1\}$ |
| $\{P_3\}$ | by | $[\bar{r}_1, \bar{r}_2, \bar{r}_3]^T$ | $\rightarrow$ | $\{P_1\}$ |
| $\{P_3\}$ | by | $[r_1, \bar{r}_2, r_3]^T$ | $\rightarrow$ | $\{P_1\}$ |
| $\{P_3\}$ | by | $[\bar{r}_1, r_2, r_3]^T$ | $\rightarrow$ | $\{P_1\}$ |
| $\{P_3\}$ | by | $[r_1, \bar{r}_2, r_3]^T$ | $\rightarrow$ | $\{P_1\}$ |
| $\{P_1 P_3\}$ | by | $[\bar{r}_1, \bar{r}_2, r_3]^T$ | $\rightarrow$ | $\{P_1\}$ |
| $\{P_1 P_3\}$ | by | $[\bar{r}_1, r_2, r_3]^T$ | $\rightarrow$ | $\{P_1\}$ |

The expression of $M_{\{1\}(k+1)}$, at time (k+1), is there-

fore:

$$\begin{aligned}
M_{\{1\}(k+1)} &= (\bar{r}_1\bar{r}_2\bar{r}_3 + \bar{r}_1 r_2 \bar{r}_3 \\
&+ \bar{r}_1\bar{r}_2 r_3 + \bar{r}_1 r_2 r_3).M_{\{1\}(k)} \\
&+ (\bar{r}_1\bar{r}_2 r_3 + r_1\bar{r}_2 r_3 \\
&+ \bar{r}_1 r_2 r_3 + r_1 r_2 r_3).M_{\{3\}(k)} \\
&+ (\bar{r}_1\bar{r}_2 r_3 + \bar{r}_1 r_2 r_3).M_{\{1,3\}(k)}
\end{aligned}$$

By factorisation of this expression, it is obtained:

$$M_{\{1\}(k+1)} = \bar{r}_1 M_{\{1\}(k)} + r_3 M_{\{3\}(k)} + \bar{r}_1 r_3 M_{\{1,3\}(k)}$$

In the same manner, the expression of the mass $m_Y$ of all the subsets of places $Y$ of $2^\Omega$ is determined. The following equations are obtained:

$$\begin{aligned}
M_{\{1\}(k+1)} &= \bar{r}_1 M_{\{1\}(k)} + r_3 M_{\{3\}(k)} \\
&+ \bar{r}_1 r_3 M_{\{1,3\}(k)} \\
M_{\{2\}(k+1)} &= \bar{r}_2 M_{\{2\}(k)} + r_1 M_{\{1\}(k)} \\
&+ \bar{r}_2 r_1 M_{\{1,2\}(k)} \\
M_{\{3\}(k+1)} &= \bar{r}_3 M_{\{3\}(k)} + r_2 M_{\{2\}(k)} \\
&+ \bar{r}_3 r_2 M_{\{2,3\}(k)} \\
M_{\{1,2\}(k+1)} &= \bar{r}_1\bar{r}_2 M_{\{1,2\}(k)} + r_1 r_3 M_{\{1,3\}(k)} \\
&+ \bar{r}_2 r_3 (M_{\{2,3\}(k)} + M_{\{1,2,3\}(k)}) \\
M_{\{1,3\}(k+1)} &= \bar{r}_1\bar{r}_3 M_{\{1,3\}(k)} + r_2 r_3 M_{\{2,3\}(k)} \\
&+ \bar{r}_1 r_2 (M_{\{1,2\}(k)} + M_{\{1,2,3\}(k)}) \\
M_{\{2,3\}(k+1)} &= \bar{r}_2\bar{r}_3 M_{\{2,3\}(k)} + r_1 r_2 M_{\{1,2\}(k)} \\
&+ \bar{r}_3 r_1 (M_{\{1,3\}(k)} + M_{\{1,2,3\}(k)}) \\
M_{\{1,2,3\}(k+1)} &= (\bar{r}_1\bar{r}_2\bar{r}_3 + r_1 r_2 r_3) M_{\{1,2,3\}(k)}
\end{aligned}$$

If the initial state of the system is given by the vector $\mathcal{M}_{(0)} = [0, 0, 0, 0, 0, 0, 1]$, (unknown initial state, $M_{\{1,2,3\}} = 1$ and $M_X = 0, \forall\ X \neq \{1,2,3\}$), and according to the sequence of the receptivities taken from the system, the new state becomes more precise. For example, if $R_{(1)} = [0,1,0]$, the new marking of the system is $\mathcal{M}_{(1)} = [0, 0, 0, 0, 1, 0, 0]$, and the mass $M_{\{1,3\}}$ of $X = \{P_1, P_3\}$ becomes 1; it is certain not to be in the place $P_2$.

### 4.2  General property

It has been shown in section 2.2, that the most important propriety in this type of net is the *p-invariant* as it is presented in equation 2. The new mark vector $\mathcal{M}_{(k)}$ is defined such as a distribution of masses on the set $2^\Omega$, so it is assumed that at time $(k)$, the p-invariant is equal to one:

$$\sum_{X \in 2^\Omega} M(X)_{(k)} = 1$$

The transformation proposed before, leads to a new vector $\mathcal{M}_{(k+1)}$. It is shown below that the p-invariant is still equal to one.

In fact, when $M(X)$ is equal to one, that means all the places $P_i \in X$ are possible (but not sure). In order to find the following state, the algorithm consists of giving a mark equal to one to each place in $X$.

$$\forall i/P_i \in X, M(\{P_i\}) = 1$$

So the usual evolution equation (1) of Petri net is applied. Because the initial mass is not equal to 1, the final mass can be different that 1. All the places with mass different to zero are then possible, but it is impossible to determine if one is more possible than an other. So, all the masses are given to the union of these places. By using this method, it is not needed to give at priori probability on each place $P_i$ of $X$, contrary in using the theory of probability. A possible element $X$ of $2^\Omega$ gives only one possible element $Y$ of $2^\Omega$ such as $M(Y) = 1$ after the evolution of the net. So, the sum of the masses after evolution is equal to one:

$$\sum_{X \in 2^\Omega} M(X)_{(k+1)} = 1$$

In the general case, the global evolution equations can not be easily deduced. In fact, the number of possible configuration is equal to $(2^{n-1} * 2^n)$ for a single net with $n$ places and $n$ transitions. If the initial mass is given for a single place $P_i$, the resolution is simplified and corresponds to a classic Petri net:

$$\left. \begin{array}{l} \{P_i\} \xrightarrow{\bar{r}_i} \\ \{P_{i-1}\} \xrightarrow{r_{i-1}} \\ \{P_{i-1}, P_i\} \xrightarrow{r_{i-1}\bar{r}_i} \end{array} \right\} \{P_i\}$$

The logical equation that computes the mass of an element set $\{P_i\}$ at the time $(k+1)$ is given by:

$$\begin{aligned}
M_{\{i\}(k+1)} &= \bar{r}_i * M_{\{i\}(k)} + r_{i-1} * M_{\{i-1\}(k)} \\
&+ r_{i-1} * \bar{r}_i * M_{\{i-1,i\}(k)}
\end{aligned}$$

But in case of unknown initial place, the general equations cannot be easily written. The more the number of places in the system increases, the more the number of possible places combinations increases. A possible solution is to decompose the net in subnets, the state of each subnet depends only on one transition $t_j$ [4].

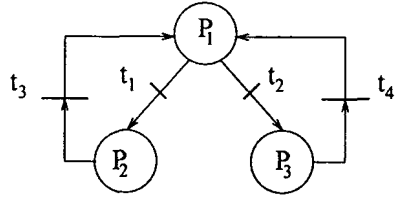

Figure 2: System with conflict

## 5 Petri net with conflict

The two transitions $t_1$ and $t_2$ are in **conflict** for a marking $M$ if and only if there is a place $P$ such that:

- $M(P) < Pre(P,t_1) + Pre(P,t_2)$
- $Pre(P,t_1) = 1$ and $Pre(P,t_2) = 1$

The usual solution in classic Petri net is to forbid a conflict where the two transitions are simultaneously true. This is equivalent to the relation $(r_1$ and $r_2) \neq 1$: $r_1 * r_2 \leq 1$, so only one token is in the net.

In the example represented (figure 2), the problem comes with the two transitions after $P_1$. The classic Petri net leads to the following equations:

$$M(\{P_2\})_{(k+1)} = \bar{r}_3 * M(\{P_2\})_{(k)} + r_1 * M(\{P_1\})_{(k)}$$

$$M(\{P_3\})_{(k+1)} = \bar{r}_4 * M(\{P_3\})_{(k)} + r_2 * M(\{P_1\})_{(k)}$$

If $r_1 = 1$, and simultaneously $r_2 = 1$ and $M(\{P_1\})_{(k)} = 1$, then $M(\{P_2\})_{(k+1)} = 1$, $M(\{P_3\})_{(k+1)} = 1$ and the hypothese of the p-invariant becomes false ($\sum_{i=1}^{3} M(\{P_i\}) = 2$). The two matrix $Pre$ and $Post$ for this type of Petri net are :

$$Pre = \begin{array}{c} \phantom{Pre =} \begin{array}{cccc} t_1 & t_2 & t_3 & t_4 \end{array} \\ \left[ \begin{array}{cccc} 1 & 1 & 0 & 0 \\ 0 & 0 & 1 & 0 \\ 0 & 0 & 0 & 1 \end{array} \right] \begin{array}{c} P_1 \\ P_2 \\ P_3 \end{array} \end{array}$$

$$Post = \begin{array}{c} \phantom{Post =} \begin{array}{cccc} t_1 & t_2 & t_3 & t_4 \end{array} \\ \left[ \begin{array}{cccc} 0 & 0 & 1 & 1 \\ 1 & 0 & 0 & 0 \\ 0 & 1 & 0 & 0 \end{array} \right] \begin{array}{c} P_1 \\ P_2 \\ P_3 \end{array} \end{array}$$

In case of conflict, some lines of these two matrix have several 1. Using the constraint on the transitions $t_1$ and $t_2$ means the token in the place $P_1$ can leave to go in the place $P_2$, in the place $P_3$, or it can remain in its initial place $P_1$, but it cannot moved into the two places $P_2$ and $P_3$ simultaneously. And therefore the two transitions $t_1$ and $t_2$ can not be fired at the same time because of this constraint.

In general case with the new mark $\mathcal{M}$, the problem is the same. It is assumed at the time $(k)$, that the mark of $X \in 2^\Omega$ is equal to one $(M(X)_{(k)} = 1)$. At the time $(k+1)$ comes a new vector of receptivity $R_{(k+1)}$. The transformation of $X$, the set $Y \in 2^\Omega$, must be determined. The same method, described in section 4 is used. All the $P_i$ of $X$ are possible, and their mark is put to one. Then, the equation (1) is applied, in order to get a new mark vector. The set $Y$ is the union of all the places with a mark different to zero.

In the case of conflict, the following assumption is made: the two transitions cannot be true simultaneously. So, if the two places of the conflict are not initially in $X$, so only one of the two places could be in $Y$. But, if one or more places of the conflict are initially in $X$, the two places can be in $Y$ after the application of equation (1).

In the example of figure (2), it is assumed that, at the time $(k)$, $M_{\{1,2\}(k)} = 1$. At the time $(k+1)$, only $r_2 = 1$, so the set $Y = \{P_2, P_3\}$. These two places in conflict can be simultaneously possible.

A Petri net with a conflict is not a problem in case of general resolution, if the same constraint on receptivities as the classic Petri net are assumed.

## 6 Conclusion and prospectives

Here, a method is proposed to deal with the initial state uncertainty of the Petri net. This method is based on the model of uncertainty from Dempster and Shafer. As in a classic Petri net, the mark of a place $P_i$ can be taken as the "probability" to be in this place. The generalized mark $\mathcal{M}$ of the Petri net corresponds to the "probability" to be in a set of places $P_i$. The equations of evolution are done, in a simple Petri net, and in a Petri net with conflict.

The first and most important problem comes with the number of operations needed to compute the expression of all marks. This calculation is long and complex. Indeed, for a system of $n$ places, the set $2^\Omega$ had $2^n - 1$ elements, and for $n$ transitions, there are $2^n$ possible combinations. The resolution of the system leads to the study of $(2^n - 1) * 2^n$ different cases. More, the equations are linked to the global structure of the net. The solution can not be found in the general case. Nevertheless, this method solves the problem when the state of the system is unknown, and verifies the principle of the mass conservation. It allows to take a decision at any time. A possible solution to this problem is to take each transitions as a source of information, and then to decompose the net in a set of subnets.

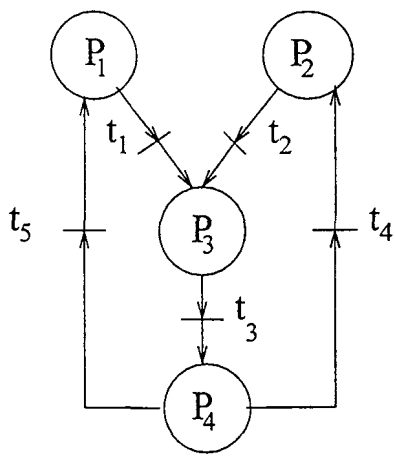

Figure 3: System with synchronization

In this paper, the problem of **synchronisation** in a Petri net such as has presented in the figure (3), has be not studied. The problem is that there are two tokens in the net. A possible solution can be the same such as in the classic Petri net: considering the net such as two independant nets.

Finally, the problem of data uncertainty on the inputs will be studied in future works. This leads to take a value in the interval $[0,1]$, instead of $\{0,1\}$ for the receptivity, [6], [7]. Some works have to be done in the case of fuzzy Petri nets [8], [9], the method must be applied to these Petri nets.